\documentclass[11pt]{article}

\usepackage[preprint]{acl}
\newcommand{\codeurl}{https://github.com/sileod/modal-semantics-reasoning}
\newcommand{\dataurl}{https://huggingface.co/datasets/sileod/modal-semantics-reasoning}
\usepackage{times}
\usepackage{latexsym}
\usepackage[T1]{fontenc}
\usepackage[utf8]{inputenc}
\usepackage{microtype}
\usepackage{booktabs}
\usepackage{float}
\usepackage{amsmath}
\usepackage{xcolor}
\usepackage{tikz}
\usepackage{graphicx}
\usetikzlibrary{arrows.meta,calc,positioning,shapes.geometric}
\usepackage[subtle]{savetrees}

\newcommand{\ci}[3]{#1\,{\tiny[#2, #3]}}
\newcommand{\best}[1]{\textbf{#1}}

\title{Same Formulas, Different Semantics:\\
Do Language Models Follow Modal Logic Specifications?}

\author{ Rémi Andrieu \and Damien Sileo\\
Univ. Lille, Inria, CNRS, Centrale Lille, UMR 9189 - CRIStAL, F-59000 Lille, France \\
\texttt{damien.sileo@inria.fr}}

\begin{document}
\maketitle
\begin{abstract}
Reasoning about necessity and possibility depends on assumptions about
accessibility between worlds and about which objects exist at each one. The
same inference may therefore hold under one modal system and fail under
another. Evaluating language models on such problems requires testing whether
their judgments follow the stated semantics rather than a familiar logic. We
construct paired modal problems with identical premises and conjecture but
different frame or domain conditions; automated reasoning verifies opposite
labels. A balanced core prevents the semantic
condition alone from revealing the answer. On this core, four of five recent
models perform below the condition-only baseline under direct
prompting. Yet enabling reasoning mode raises DeepSeek V4 Flash from
4.4\% to 88.1\% on unchanged prompts. Following stipulated modal semantics
thus depends strongly on inference mode as well as model identity. When frame
conditions are omitted, models often agree but fit different familiar logics
best. We release the formulas, oracle artifacts,
countermodels, and responses.
\end{abstract}

\section{Introduction}

Whether an inference is valid can depend on the semantics rather than on its
surface form. This dependence is especially pronounced for necessity and
possibility. For example, ``if $p$ is necessary, then $p$ is possible'' holds
when every world accesses another world, but fails on arbitrary Kripke frames.
Similarly, moving a quantifier across necessity can be licensed or blocked by
whether objects may appear or disappear across worlds
\citep{kripke1963semantical,barcan1946functional,fitting1998first}.
Reasoning correctly in such settings requires following the declared frame and
domain assumptions rather than silently substituting a familiar modal logic.
This matters in deontic or legal reasoning, and more generally whenever a
system must reason under stated constraints.

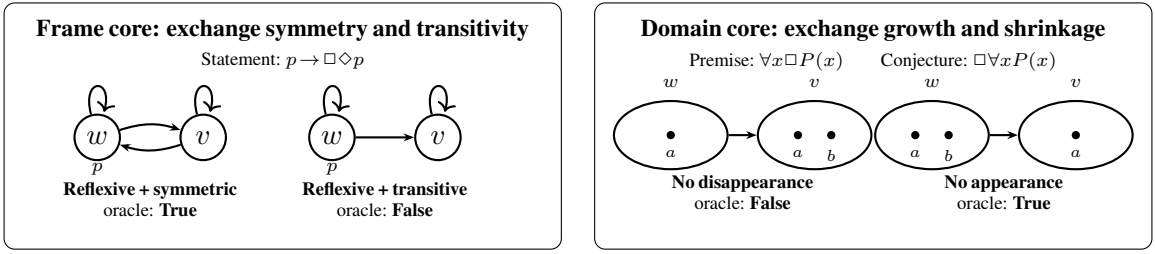
\begin{figure*}[t]
\centering
\begin{tikzpicture}[
  world/.style={circle,draw,thick,minimum size=6mm,inner sep=0pt},
  domainworld/.style={ellipse,draw,thick,minimum width=15mm,minimum height=9mm},
  edge/.style={-{Stealth[length=4pt]},thick},
  panel/.style={draw,rounded corners,inner sep=5pt},
  note/.style={font=\scriptsize,align=center},
  dot/.style={circle,fill,inner sep=1.1pt}
]
\node[panel,minimum width=.46\textwidth,minimum height=3.25cm] (frame) at (-3.9,0) {};
\node[panel,minimum width=.46\textwidth,minimum height=3.25cm] (domains) at (3.9,0) {};

\node[font=\bfseries\small] at (-3.9,1.27) {Frame core: exchange symmetry and transitivity};
\node[note] at (-3.9,.86) {Statement: $p\rightarrow\Box\Diamond p$};

\begin{scope}[shift={(-6.35,-.15)}]
  \node[world] (tw) at (0,0) {$w$};
  \node[world] (tv) at (1.4,0) {$v$};
  \draw[edge] (tw) to[bend left=18] (tv);
  \draw[edge] (tv) to[bend left=18] (tw);
  \draw[edge] (tw) edge[loop above] (tw);
  \draw[edge] (tv) edge[loop above] (tv);
  \node[note] at (0,-.42) {$p$};
  \node[note] at (.7,-.82) {\textbf{Reflexive + symmetric}\\oracle: \textbf{True}};
\end{scope}
\begin{scope}[shift={(-3.25,-.15)}]
  \node[world] (sw) at (0,0) {$w$};
  \node[world] (sv) at (1.4,0) {$v$};
  \draw[edge] (sw) -- (sv);
  \draw[edge] (sw) edge[loop above] (sw);
  \draw[edge] (sv) edge[loop above] (sv);
  \node[note] at (0,-.42) {$p$};
  \node[note] at (.7,-.82) {\textbf{Reflexive + transitive}\\oracle: \textbf{False}};
\end{scope}

\node[font=\bfseries\small] at (3.9,1.27) {Domain core: exchange growth and shrinkage};
\node[note] at (3.9,.86)
  {Premise: $\forall x\,\Box P(x)$\qquad Conjecture: $\Box\forall x\,P(x)$};

\begin{scope}[shift={(1.225,-.12)}]
  \node[domainworld] (vw) at (0,0) {};
  \node[domainworld] (vv) at (1.9,0) {};
  \draw[edge] (vw) -- (vv);
  \node[note,above=1pt of vw] {$w$};
  \node[note,above=1pt of vv] {$v$};
  \node[dot,label=below:{\tiny $a$}] at (vw) {};
  \node[dot,label=below:{\tiny $a$}] at ($(vv)+(-.22,0)$) {};
  \node[dot,label=below:{\tiny $b$}] at ($(vv)+(.22,0)$) {};
  \node[note] at (.95,-.75) {\textbf{No disappearance}\\oracle: \textbf{False}};
\end{scope}
\begin{scope}[shift={(4.675,-.12)}]
  \node[domainworld] (cw) at (0,0) {};
  \node[domainworld] (cv) at (1.9,0) {};
  \draw[edge] (cw) -- (cv);
  \node[note,above=1pt of cw] {$w$};
  \node[note,above=1pt of cv] {$v$};
  \node[dot,label=below:{\tiny $a$}] at ($(cw)+(-.22,0)$) {};
  \node[dot,label=below:{\tiny $b$}] at ($(cw)+(.22,0)$) {};
  \node[dot,label=below:{\tiny $a$}] at (cv) {};
  \node[note] at (.95,-.75) {\textbf{No appearance}\\oracle: \textbf{True}};
\end{scope}
\end{tikzpicture}
\caption{The two balanced-core contrasts. The problem formula is fixed within
each pair; only one explicit rule changes. Each rule occurs equally
often with either oracle label, so strict accuracy requires reading the
formula rather than mapping a semantic condition to an answer.}
\label{fig:setup}
\end{figure*}

Most NLP reasoning benchmarks instead assume a fixed background logic and vary
the facts, rules, or proof depth
\citep{tafjord2021proofwriter,han2022folio,parmar2024logicbench}. A model can
therefore perform well by learning the benchmark's dominant inference regime.
Modal-reasoning evaluations broaden the class of inferences, but generally
still ask whether a model solves individual problems under one intended
semantics
\citep{holliday2024conditionalmodalreasoninglarge,Li2025ModalLogicBenchUM}.
They do not directly test whether a model's judgments change when the semantic
specification changes.
We study this specification sensitivity by holding the object-level problem
fixed while varying its semantics. Each problem pairs identical premises and
conjecture under two specifications that differ in one frame or domain
condition, with an automated-reasoning oracle verifying opposite labels.
Success therefore requires tracking the stated semantics rather than applying
one fixed modal logic. We separately omit frame specifications to examine the
logics models favor when the prompt leaves them unconstrained.
The design combines three complementary features. First, lexical content and
formula difficulty are fixed within each pair. Second, the broad nested set captures the general effects of stronger modal conditions across many formulas but admits a condition shortcut; our primary balanced core removes that shortcut by balancing each condition across labels.
Third, failures are easy to inspect because the semantic intervention is
explicit. Together, these features distinguish reasoning that adapts to the
specification from success under a familiar default logic.
Figure~\ref{fig:setup} summarizes the evaluation design.
\textbf{Resources:}\enspace
\href{\codeurl}{\raisebox{-0.15ex}{\includegraphics[height=0.9em]{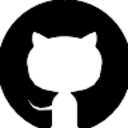}}\,Code}
\quad
\href{\dataurl}{\raisebox{-0.15ex}{\includegraphics[height=0.9em]{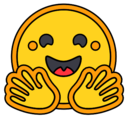}}\,Data}.

\section{Related Work}

ProofWriter, FOLIO, LogicNLI, and LogicBench evaluate deduction under a fixed
intended logic
\citep{tafjord2021proofwriter,han2022folio,tian-etal-2021-diagnosing,
parmar2024logicbench}. Modal evaluations include controlled syllogisms,
dynamic epistemic reasoning, and other fixed-semantics problems
\citep{wang-shi-2025-logical,sileo-lernould-2023-mindgames,
holliday2024conditionalmodalreasoninglarge,Li2025ModalLogicBenchUM}.
QMLTP, the interoperable non-classical TPTP format, and embedding-based theorem
proving provide infrastructure for quantified modal reasoning
\citep{raths2012qmltp,steen2025tptp,steen2024solvingqml}. We use that formal
infrastructure as an oracle rather than proposing a new logic or prover. Our diagnostic tests whether models follow modal specifications by holding
the linguistic problem fixed, changing one declared model-theoretic condition,
and requiring both judgments to be correct.
Methodologically, this resembles contrast sets and SpaceNLI's pattern accuracy
\citep{gardner-etal-2020-contrast,abzianidze-etal-2023-spacenli}. Unlike a
linguistic perturbation, however, our premises and conjecture remain unchanged:
the intervention is in the declared model-theoretic semantics.

To our knowledge, this is the broadest controlled evaluation of whether LLMs
follow modal semantics, covering five frame-property contrasts and three
first-order domain contrasts.

\section{Task formulation}

We evaluate each problem under two specifications:
\[
((S_a,P,C),y_a),\quad ((S_b,P,C),y_b),
\]
where \(S\) gives the semantics, \(P\) the possibly empty premise set,
\(C\) the conjecture, and \(y\in\{\mathrm{true},\mathrm{false}\}\).
Retained pairs satisfy
\[
P_a=P_b,\quad C_a=C_b,\quad y_a\neq y_b,
\]
and \(S_a,S_b\) differ in one frame or domain condition.

When $P$ is empty, the prompt asks whether $C$ is valid under $S$. Otherwise it
asks whether $C$ follows from the premises. This distinction avoids presenting
formula-validity problems as inference from an artificial empty premise list.

\paragraph{Frame semantics.}
We use the familiar systems K, D, T, B, S4, and S5, while storing their explicit
frame properties. The controlled contrasts add seriality (every world accesses
some world), reflexivity (every world accesses itself), symmetry (accessibility
holds in both directions), or transitivity (two accessibility steps compose):
K--D, K--T, T--B, T--S4, and B--S5, respectively. Frame problems are
propositional, so domain and name semantics cannot affect the answer.

\paragraph{Domain semantics.}
Domains are varying, cumulative, decreasing, or constant. Cumulative domains
prevent objects from disappearing along accessibility; decreasing domains
prevent new objects from appearing; constant domains impose both constraints.
All domain problems use serial frames (system D), variables and predicates, and
no constants or functions. Their formulas contain quantifier--modality
alternation, such as $\Box\forall x\,P(x)$ versus
$\forall x\,\Box P(x)$.

\paragraph{Balanced non-nested core.}
Because nested systems make the stronger condition predictive of validity, we add 160 non-nested pairs. B versus S4 exchanges symmetry and transitivity on a reflexive base; cumulative versus decreasing exchanges growth-only and shrink-only domains. Crossing these contrasts with two flip directions and two task types yields eight 20-pair cells. Thus each condition occurs equally often with each label, and the formula determines the flip.

\paragraph{Controlled English.}
The renderer is deterministic and directly states the relevant semantics.
Prompts spell out rules but withhold conventional system names; B, S4,
cumulative, and decreasing are table shorthand only.
Propositions and predicates use ordinary neutral vocabulary, such as ``the
signal is active'' and ``the object is registered.'' Nested modal scope is
expressed relative to the current world and then ``that world.'' The primary
protocol requests only a binary Yes/No judgment.

\begin{table*}[t]
\centering
\scriptsize
\setlength{\tabcolsep}{4pt}
\begin{tabular}{p{0.47\textwidth}p{0.47\textwidth}}
\toprule
\textbf{Frame: reflexive + symmetric (True)} &
\textbf{Frame: reflexive + transitive (False)} \\
\midrule
\textit{Semantic rule:} The accessibility relation is reflexive and symmetric.
&
\textit{Semantic rule:} The accessibility relation is reflexive and transitive.
\\[2pt]
\multicolumn{2}{p{0.96\textwidth}}{\textit{Premise:} At the current world, the following holds: (the signal is active and the alarm is sounding).}\\
\multicolumn{2}{p{0.96\textwidth}}{\textit{Conjecture:} At every world accessible from the current world, there is a world accessible from that world where the following holds: (the signal is active and the alarm is sounding).}\\
\midrule
\textbf{Domain: no disappearance (False)} &
\textbf{Domain: no appearance (True)} \\
\midrule
\textit{Semantic rule:} Objects cannot disappear when moving to an accessible world.
&
\textit{Semantic rule:} New objects cannot appear when moving to an accessible world.
\\[2pt]
\multicolumn{2}{p{0.96\textwidth}}{\textit{Premise:} Every object existing at the current world has status ((registered or approved) and verified) at every accessible world.}\\
\multicolumn{2}{p{0.96\textwidth}}{\textit{Conjecture:} At every accessible world, every object existing there has status ((registered or approved) and verified).}\\
\bottomrule
\end{tabular}
\caption{Examples from the balanced core. Premise and conjecture are
identical within each pair, and prompts contain only the rules shown. The first
favors symmetry and the second shrink-only domains; reverse directions are
equally represented.}
\label{tab:examples}
\end{table*}

\section{Construction and Oracle}

Candidate families intentionally target one contrast instead of sampling the
Cartesian product of formulas and semantics. We generate both compact validity
schemas and premise-bearing versions. Structural checks reject first-order
constructs on the frame axis, constants on the domain axis, unintended semantic
changes, duplicate canonical formulas, and excessive depth.
Separate from this core, the broad nested set contains 800 pairs: 80 for each
of five Frame contrasts;
varying--cumulative and varying--decreasing have 134 Domain pairs each, and
cumulative--constant has 132. Validity and premise-bearing inference each
contribute 400 pairs.
Premise ablation removes the premises, yielding 395 premise-dependent pairs,
400 conjecture-only pairs, and five unresolved ablations
(Table~\ref{tab:premise-analysis} in the Appendix).

Problems are serialized in non-classical TPTP, a machine-readable logic syntax,
and translated to higher-order logic using the LET embedding toolchain
\citep{steen2022let}. Vampire and Leo-III return standard SZS proof or
countermodel statuses
\citep{kovacs2013vampire,steen2021leo3}. We never infer invalidity from failure
to prove validity. Exact source problems, translations, commands, versions,
runtimes, exit codes, and output hashes are retained.

Prover coverage is asymmetric: Leo-III primarily proves valid sides, whereas
Vampire primarily supplies countermodels for invalid sides. We therefore report
dual-prover agreement separately from single-prover resolutions and reject
every conflict or unresolved side. Of 1,600 accepted sides, 727 have dual
agreement and 873 have one decisive result with no contradiction; 15 timed-out
candidates were discarded. In the balanced core, every invalid side
additionally has a two- or three-world countermodel checked by an independent
Kripke evaluator; 136 of 160 valid sides have dual ATP agreement and the
remaining 24 have one proof.

\begin{table*}[t]
\centering
\small
\setlength{\tabcolsep}{1pt}
\begin{tabular}{lrrrrr}
\toprule
Model & B$\leftrightarrow$S4$\uparrow$ & Cum.$\leftrightarrow$Dec.$\uparrow$ & Mean$\uparrow$ & Parsed acc.$\uparrow$ & Pair parse$\uparrow$ \\
\midrule
DeepSeek V4 Flash & \ci{1.2}{0.2}{6.7} & \ci{7.5}{3.5}{15.4} & \ci{4.4}{2.1}{8.8} & \ci{4.4}{2.1}{8.8} & \ci{99.4}{96.5}{99.9} \\
DeepSeek V4 Pro & \ci{0.0}{0.0}{4.6} & \ci{5.0}{2.0}{12.2} & \ci{2.5}{1.0}{6.3} & \ci{2.5}{1.0}{6.3} & \ci{100.0}{97.7}{100.0} \\
GPT-5.6 Luna & \ci{7.5}{3.5}{15.4} & \ci{35.0}{25.5}{45.9} & \ci{21.2}{15.6}{28.2} & \ci{21.2}{15.6}{28.2} & \ci{100.0}{97.7}{100.0} \\
GPT-5.6 Terra & \ci{3.8}{1.3}{10.5} & \ci{46.2}{35.7}{57.1} & \ci{25.0}{18.9}{32.2} & \ci{25.0}{18.9}{32.2} & \ci{100.0}{97.7}{100.0} \\
Claude Sonnet 5 & \ci{56.2}{45.3}{66.6} & \ci{73.8}{63.2}{82.1} & \ci{65.0}{57.3}{72.0} & \best{\ci{89.7}{82.8}{94.0}} & \ci{72.5}{65.1}{78.8} \\
\midrule
DeepSeek V4 Flash (high) & \best{\ci{91.2}{83.0}{95.7}} & \best{\ci{85.0}{75.6}{91.2}} & \best{\ci{88.1}{82.2}{92.3}} & \ci{88.7}{82.8}{92.7} & \ci{99.4}{96.5}{99.9} \\
\midrule
Condition-only baseline & 50.0 & 50.0 & 50.0 & -- & -- \\
\bottomrule
\end{tabular}
\caption{Balanced non-nested core. Each of the same two conditions
occurs under both labels within each contrast; half the formulas favor either
condition. Thus a condition-only strategy reaches 50\% strict accuracy, while
independent random answers reach 25\%. Every row uses all 160 pairs, and prompts
show rules rather than conventional system names. Malformed responses are
wrong; Parsed acc. conditions on both answers parsing, while Pair parse is the
fraction of pairs with two parsed answers. Intervals are 95\% Wilson.}
\label{tab:balanced-core}
\end{table*}

\section{Experiments}

\paragraph{Models and protocol.}
We evaluate dated endpoints for DeepSeek V4 Flash and Pro, GPT-5.6 Luna and
Terra, and Claude Sonnet 5 through OpenRouter. Exact identifiers and request
parameters live in a versioned configuration. Every model receives all 800
pairs under direct inference (without reasoning mode), at temperature zero
with one response per side. Malformed answers are incorrect without repair; raw
responses are stored unchanged before scoring. We additionally evaluate all 400 Frame
pairs with high reasoning mode for Flash and medium reasoning for
Luna, holding prompts fixed; Flash also receives all 400 Domain pairs at high
effort. A matched 50-pair study compares three representations using the main-table
Terra endpoint. All five direct models and Flash high reasoning
receive the complete 160-pair core;
exact maximum output tokens are recorded in the Appendix and configuration.

\paragraph{Metrics.}
Side accuracy scores individual specifications; strict pair accuracy requires both judgments in a pair to be correct
\[
 \frac{1}{N}\sum_i
 \mathbf{1}[\hat y_{i,a}=y_{i,a}\land \hat y_{i,b}=y_{i,b}].
\]
Independent random answers score 25\% in expectation, while a constant answerer
or a model applying the same fixed semantics to both sides scores 0\%. On the
balanced core, a strategy that maps each stated condition to its optimal label
without reading the formula scores 50\% strict pair accuracy. Exceeding 50\%
therefore requires using the formula to determine which condition validates it.
Answer-change rate, $\Pr(\hat y_a\ne\hat y_b)$, separates ignoring an
intervention from reacting to it; we also inspect correctness conditional on a
change and the four correct/incorrect side outcomes.
We report 95\% Wilson intervals for binary accuracies and parse rates; this
avoids degenerate zero-width intervals for zero-success cells. The
unweighted cross-axis mean uses a stratified pair bootstrap. We additionally
report validity versus NLI performance and use premise dependence only as a
compact diagnostic.

\paragraph{Semantic affinity.}
We omit frame specifications from all 400 Frame problems and query opposite
polarities to control answer-label bias. The resulting vector is compared with
K, D, T, B, S4, and S5, weighting contrasts equally and retaining ties.

\paragraph{Representation sensitivity.}
On a deterministic 50-pair Frame subset (10 per contrast), we compare named
English, relational definitions, and TPTP with all other fields matched.
\section{Results and Analysis}

\paragraph{A failure of semantic control.}
The balanced core reveals more than difficult formulas. Four of five models
score below the 50\% condition-only baseline under direct prompting, ranging
from 2.5\% to 25.0\%; only Sonnet exceeds it at 65.0\%. Because the formula is
fixed within each pair, changing only the stipulated semantics often fails to
change the judgment. Accuracy can hide this: one side may be correct even
when the model ignores the contrast determining the other.

\paragraph{Defaults are real, but not decisive.}
Without specifications, models exhibit coherent affinities with familiar logics
such as K or T. These defaults explain some agreement on underspecified
problems, but do not reliably predict explicit errors. Models must do more than
start from the right logic: they must suspend a familiar inference regime and
let the declared model class govern the current problem.

\paragraph{Reasoning can restore control.}
With unchanged prompts, DeepSeek V4 Flash rises from 4.4\% to 88.1\% on the
balanced core; the pattern also appears on the broader Frame and Domain sets,
and for Luna on Frame. This is more specific than saying that more reasoning
improves accuracy: inference-time computation changes whether the model reacts
to the semantic intervention at all. It still does not guarantee correctness.
A plausible derivation may import an unstated property, such as using
reflexivity where transitivity is required.

\paragraph{Representation is not a simple fix.}
The matched pilot also argues against a purely surface-level explanation. For
Terra, strict Frame accuracy moves from 38\% with named conditions to 6\% with
relational definitions and 44\% with TPTP; the other models show different
rankings. Representation matters, but no format consistently repairs semantic
control. Formal syntax changes which errors appear without removing the need to
follow the stipulated model class.

\section{Conclusion}

These experiments separate modal knowledge from semantic control. A model may
exhibit a coherent default logic yet fail to let a local specification govern
its answer; additional computation can restore that sensitivity without
guaranteeing valid intermediate steps. Fixed-semantics benchmarks may therefore
overstate robustness.

\section*{Limitations}
The benchmark uses controlled English and covers only single-modality frame and domain semantics. We exclude flexible names because the prover portfolio did not reliably resolve their countermodels, and multi-agent cases to keep interventions focused. Domain contrasts use serial frames, so transfer to other frame classes remains open. Labels inherit the LET embedding and prover assumptions, and success on synthetic formulas does not establish robust modal reasoning in natural discourse. The small representation study leaves room for model- and contrast-specific effects; broader paraphrase and few-shot tests remain future work. We test compliance with explicit semantics, not difficulty for untrained humans, so we do not compare against a human baseline. API reasoning levels are neither transparent nor calibrated across vendors and are treated only as within-model conditions. The nested set leaks label information through condition names, so formula-sensitive claims rest on the balanced core. Finally, each API condition uses one sample and dated endpoints may still change. Transport failures are retried twice; successful empty responses are neither repaired nor retried and remain reflected in both reported and parse-conditional scores.
\bibliography{custom}

\begin{thebibliography}{19}
\providecommand{\natexlab}[1]{#1}

\bibitem[{Abzianidze et~al.(2023)Abzianidze, Zwarts, and
  Winter}]{abzianidze-etal-2023-spacenli}
Lasha Abzianidze, Joost Zwarts, and Yoad Winter. 2023.
\newblock \href {https://aclanthology.org/2023.naloma-1.2/} {{SpaceNLI}:
  Evaluating the consistency of predicting inferences in space}.
\newblock In \emph{Proceedings of the 4th Natural Logic Meets Machine Learning
  Workshop}, pages 12--24. Association for Computational Linguistics.

\bibitem[{Barcan(1946)}]{barcan1946functional}
Ruth~C. Barcan. 1946.
\newblock \href {https://doi.org/10.2307/2269159} {A functional calculus of
  first order based on strict implication}.
\newblock \emph{Journal of Symbolic Logic}, 11(1):1--16.

\bibitem[{Fitting and Mendelsohn(1998)}]{fitting1998first}
Melvin Fitting and Richard~L. Mendelsohn. 1998.
\newblock \emph{First-Order Modal Logic}.
\newblock Kluwer Academic Publishers.

\bibitem[{Gardner et~al.(2020)Gardner, Artzi, Basmov, Berant, Bogin, Chen,
  Dasigi, Dua, Elazar, Gottumukkala, Gupta, Hajishirzi, Ilharco, Khashabi, Lin,
  Liu, Liu, Mulcaire, Ning, Singh, Smith, Subramanian, Tsarfaty, Wallace,
  Zhang, and Zhou}]{gardner-etal-2020-contrast}
Matt Gardner, Yoav Artzi, Victoria Basmov, Jonathan Berant, Ben Bogin, Sihao
  Chen, Pradeep Dasigi, Dheeru Dua, Yanai Elazar, Ananth Gottumukkala, Nitish
  Gupta, Hannaneh Hajishirzi, Gabriel Ilharco, Daniel Khashabi, Kevin Lin,
  Jiangming Liu, Nelson~F. Liu, Phoebe Mulcaire, Qiang Ning, and 7 others.
  2020.
\newblock \href {https://doi.org/10.18653/v1/2020.findings-emnlp.117}
  {Evaluating models' local decision boundaries via contrast sets}.
\newblock In \emph{Findings of the Association for Computational Linguistics:
  EMNLP 2020}, pages 1307--1323. Association for Computational Linguistics.

\bibitem[{Han et~al.(2022)Han, Schoelkopf, Zhao, Qi, Riddell, Zhou, Coady,
  Peng, Qiao, Benson, Sun, Wardle-Solano, Szabo, Zubova, Burtell, Fan, Liu,
  Wong, Sailor, Ni, Nan, Kasai, Yu, Zhang, Fabbri, Kryscinski, Yavuz, Liu, Lin,
  Joty, Zhou, Xiong, Ying, Cohan, and Radev}]{han2022folio}
Simeng Han, Hailey Schoelkopf, Yilun Zhao, Zhenting Qi, Martin Riddell, Wenfei
  Zhou, James Coady, David Peng, Yujie Qiao, Luke Benson, Lucy Sun, Alex
  Wardle-Solano, Hannah Szabo, Ekaterina Zubova, Matthew Burtell, Jonathan Fan,
  Yixin Liu, Brian Wong, Malcolm Sailor, and 16 others. 2022.
\newblock {FOLIO}: Natural language reasoning with first-order logic.
\newblock In \emph{Proceedings of the 2022 Conference on Empirical Methods in
  Natural Language Processing}.

\bibitem[{Holliday et~al.(2024)Holliday, Mandelkern, and
  Zhang}]{holliday2024conditionalmodalreasoninglarge}
Wesley~H. Holliday, Matthew Mandelkern, and Cedegao~E. Zhang. 2024.
\newblock \href {https://arxiv.org/abs/2401.17169} {Conditional and modal
  reasoning in large language models}.
\newblock \emph{Preprint}, arXiv:2401.17169.

\bibitem[{Kov{\'a}cs and Voronkov(2013)}]{kovacs2013vampire}
Laura Kov{\'a}cs and Andrei Voronkov. 2013.
\newblock \href {https://doi.org/10.1007/978-3-642-39799-8_1} {First-order
  theorem proving and {Vampire}}.
\newblock In \emph{Computer Aided Verification}, volume 8044 of \emph{Lecture
  Notes in Computer Science}, pages 1--35. Springer.

\bibitem[{Kripke(1963)}]{kripke1963semantical}
Saul~A. Kripke. 1963.
\newblock Semantical considerations on modal logic.
\newblock \emph{Acta Philosophica Fennica}, 16:83--94.

\bibitem[{Li et~al.(2025)Li, Liu, Zhang, Jiang, and
  Chen}]{Li2025ModalLogicBenchUM}
Xianglong Li, Yu~Liu, Botao Zhang, Mingjing Jiang, and Yunfei Chen. 2025.
\newblock \href {https://api.semanticscholar.org/CorpusID:280409061}
  {Modallogicbench: Unveiling modal logic reasoning abilities of large language
  models}.
\newblock In \emph{International Conference on Intelligent Computing}.

\bibitem[{Parmar et~al.(2024)Parmar, Patel, Varshney, Nakamura, Luo, Mashetty,
  Mitra, and Baral}]{parmar2024logicbench}
Mihir Parmar, Nisarg Patel, Neeraj Varshney, Mutsumi Nakamura, Man Luo, Santosh
  Mashetty, Arindam Mitra, and Chitta Baral. 2024.
\newblock {LogicBench}: Towards systematic evaluation of logical reasoning
  ability of large language models.
\newblock In \emph{Proceedings of the 62nd Annual Meeting of the Association
  for Computational Linguistics}.

\bibitem[{Raths and Otten(2012)}]{raths2012qmltp}
Thomas Raths and Jens Otten. 2012.
\newblock \href {https://doi.org/10.1007/978-3-642-31365-3_35} {The {QMLTP}
  problem library for first-order modal logics}.
\newblock In \emph{Automated Reasoning}, volume 7364 of \emph{Lecture Notes in
  Computer Science}, pages 454--461. Springer.

\bibitem[{Sileo and Lernould(2023)}]{sileo-lernould-2023-mindgames}
Damien Sileo and Antoine Lernould. 2023.
\newblock \href {https://doi.org/10.18653/v1/2023.findings-emnlp.303}
  {{M}ind{G}ames: Targeting theory of mind in large language models with
  dynamic epistemic modal logic}.
\newblock In \emph{Findings of the Association for Computational Linguistics:
  EMNLP 2023}, pages 4570--4577. Association for Computational Linguistics.

\bibitem[{Steen(2022)}]{steen2022let}
Alexander Steen. 2022.
\newblock An extensible logic embedding tool for lightweight non-classical
  reasoning.
\newblock In \emph{Proceedings of the 8th Workshop on Practical Aspects of
  Automated Reasoning}, volume 3201 of \emph{CEUR Workshop Proceedings}.

\bibitem[{Steen and Benzm{\"u}ller(2021)}]{steen2021leo3}
Alexander Steen and Christoph Benzm{\"u}ller. 2021.
\newblock Extensional higher-order paramodulation in {Leo-III}.
\newblock \emph{Journal of Automated Reasoning}, 65:775--807.

\bibitem[{Steen and Sutcliffe(2025)}]{steen2025tptp}
Alexander Steen and Geoff Sutcliffe. 2025.
\newblock \href {https://doi.org/10.48550/arXiv.2508.09318} {{TPTP} world
  infrastructure for non-classical logics}.
\newblock \emph{Preprint}, arXiv:2508.09318.

\bibitem[{Steen et~al.(2024)Steen, Sutcliffe, and
  Benzm{\"u}ller}]{steen2024solvingqml}
Alexander Steen, Geoff Sutcliffe, and Christoph Benzm{\"u}ller. 2024.
\newblock Solving quantified modal logic problems by translation to classical
  logics.
\newblock \emph{Journal of Automated Reasoning}.
\newblock Also available as arXiv:2212.09570.

\bibitem[{Tafjord et~al.(2021)Tafjord, Dalvi, and
  Clark}]{tafjord2021proofwriter}
Oyvind Tafjord, Bhavana Dalvi, and Peter Clark. 2021.
\newblock {ProofWriter}: Generating implications, proofs, and abductive
  statements over natural language.
\newblock In \emph{Findings of the Association for Computational Linguistics:
  ACL-IJCNLP 2021}, pages 3621--3634.

\bibitem[{Tian et~al.(2021)Tian, Li, Chen, Xiao, He, and
  Jin}]{tian-etal-2021-diagnosing}
Jidong Tian, Yitian Li, Wenqing Chen, Liqiang Xiao, Hao He, and Yaohui Jin.
  2021.
\newblock \href {https://doi.org/10.18653/v1/2021.emnlp-main.303} {Diagnosing
  the first-order logical reasoning ability through {L}ogic{NLI}}.
\newblock In \emph{Proceedings of the 2021 Conference on Empirical Methods in
  Natural Language Processing}, pages 3738--3747. Association for Computational
  Linguistics.

\bibitem[{Wang and Shi(2025)}]{wang-shi-2025-logical}
Yixuan Wang and Freda Shi. 2025.
\newblock \href {https://doi.org/10.18653/v1/2025.acl-long.824} {Logical forms
  complement probability in understanding language model (and human)
  performance}.
\newblock In \emph{Proceedings of the 63rd Annual Meeting of the Association
  for Computational Linguistics (Volume 1: Long Papers)}, pages 16862--16877.
  Association for Computational Linguistics.

\end{thebibliography}

\clearpage
\appendix
\section{Formal Semantic Conventions}
Models have a nonempty set of worlds, a designated current world, and a binary
accessibility relation. Local consequence evaluates premises and conjecture at
that world. Each world has a nonempty domain $D_w$ inside a common object
universe. Quantifiers are actualist and range over $D_w$; variable assignments
retain the same object across modal evaluation. Predicate extensions are
world-relative over the common universe and unconstrained outside $D_w$;
existence guards restrict quantification. Cumulative domains satisfy
$wRv\Rightarrow D_w\subseteq D_v$; decreasing domains reverse the inclusion;
constant domains satisfy both. Frame properties have their standard
first-order definitions. Released TPTP problems set local terms and rigid
designation; benchmark formulas contain no individual constants.

For example, $\Box q\rightarrow\Box\Box q$ is valid on S4 frames but fails on
the reflexive symmetric frame with worlds $0,1,2$, reflexive edges,
$0R1,1R0,1R2,2R1$, and no $0R2$, when $q$ is true at 0 and 1 but false at 2.
The released checker verifies this and the analogous B and domain witnesses.

\section{Secondary Diagnostic Tables}
\begin{table*}[t]
\centering
\scriptsize
\setlength{\tabcolsep}{4pt}
\begin{tabular}{lrrrrr}
\toprule
Model & Frame$\uparrow$ & Domain$\uparrow$ & Mean$\uparrow$ & Side$\uparrow$ & Parse$\uparrow$ \\
\midrule
DeepSeek V4 Flash & \ci{5.8}{3.5}{8.3} & \ci{17.5}{13.3}{21.9} & \ci{11.6}{9.2}{14.1} & \ci{54.8}{52.4}{57.2} & \ci{96.2}{94.7}{97.4} \\
DeepSeek V4 Pro & \ci{5.8}{3.4}{8.3} & \ci{31.5}{26.5}{36.6} & \ci{18.6}{15.9}{21.5} & \ci{58.9}{56.4}{61.3} & \ci{100.0}{99.5}{100.0} \\
GPT-5.6 Luna & \ci{18.2}{14.4}{22.3} & \ci{58.2}{52.9}{63.8} & \ci{38.2}{34.9}{41.6} & \ci{68.8}{66.5}{71.0} & \ci{100.0}{99.5}{100.0} \\
GPT-5.6 Terra & \ci{31.0}{26.0}{36.2} & \best{\ci{94.8}{92.3}{97.1}} & \ci{62.9}{60.2}{65.7} & \ci{81.4}{79.5}{83.3} & \ci{100.0}{99.5}{100.0} \\
Claude Sonnet 5 & \best{\ci{71.0}{65.9}{76.1}} & \ci{87.2}{83.5}{90.7} & \best{\ci{79.1}{75.9}{82.2}} & \best{\ci{89.4}{87.8}{90.9}} & \ci{87.6}{85.2}{89.7} \\
\midrule
Condition-only baseline & 60.0 & 67.0 & 63.5 & 81.8 & -- \\
\bottomrule
\end{tabular}
\caption{Direct strict accuracy on the 800 nested-system pairs. Frame, Domain, and Mean
intervals resample formula-skeleton clusters; Side and Parse use Wilson
intervals. Malformed responses are incorrect. The condition-only baseline reads
only the named semantic condition, never the formula, and outputs its optimal
label; this table is therefore a broad compliance diagnostic rather than the
balanced-core result. Bold marks column-best accuracy; Parse is not ranked.}
\label{tab:main-results}
\end{table*}

\begin{table}[H]
\centering
\scriptsize
\setlength{\tabcolsep}{2.5pt}
\begin{tabular}{llrrrrr}
\toprule
Model & Axis & Both & Only A & Only B & Neither & Change \\
\midrule
DeepSeek V4 Flash & Frame & 23 & 182 & 186 & 9 & 6.4 \\
DeepSeek V4 Flash & Domain & 70 & 158 & 165 & 7 & 18.8 \\
DeepSeek V4 Pro & Frame & 23 & 189 & 187 & 1 & 6.0 \\
DeepSeek V4 Pro & Domain & 126 & 139 & 129 & 6 & 33.0 \\
GPT-5.6 Luna & Frame & 73 & 159 & 164 & 4 & 19.2 \\
GPT-5.6 Luna & Domain & 233 & 83 & 83 & 1 & 58.5 \\
GPT-5.6 Terra & Frame & 124 & 137 & 139 & 0 & 31.0 \\
GPT-5.6 Terra & Domain & 379 & 10 & 11 & 0 & 94.8 \\
Claude Sonnet 5 & Frame & 284 & 52 & 62 & 2 & 84.3 \\
Claude Sonnet 5 & Domain & 349 & 23 & 28 & 0 & 95.9 \\
\bottomrule
\end{tabular}
\caption{Pair outcome decomposition. Direct results use 400 pairs per axis.
Both/Only/Neither denote
which sides are correct; Change is the percentage of parsed pairs receiving
different answers. Overall pair parse rates with 95\% Wilson intervals are: DeepSeek V4 Flash 96.2\%~{\scriptsize[94.7, 97.4]}; DeepSeek V4 Pro 100.0\%~{\scriptsize[99.5, 100.0]}; GPT-5.6 Luna 100.0\%~{\scriptsize[99.5, 100.0]}; GPT-5.6 Terra 100.0\%~{\scriptsize[99.5, 100.0]}; Claude Sonnet 5 87.6\%~{\scriptsize[85.2, 89.7]}.}
\label{tab:pair-outcomes}
\end{table}

\begin{table}[t]
\centering
\scriptsize
\setlength{\tabcolsep}{2pt}
\resizebox{\columnwidth}{!}{\begin{tabular}{lrrrr}
\toprule
Model & Direct & Reasoning & $\Delta$ & Parse \\
\midrule
V4 Flash Frame (high) & \ci{5.8}{3.9}{8.5} & \best{\ci{92.5}{89.5}{94.7}} & \ci{+86.8}{83.2}{90.0} & 98.8 \\
V4 Flash Domain (high) & \ci{17.5}{14.1}{21.5} & \ci{90.0}{86.7}{92.6} & \ci{+72.5}{67.5}{77.2} & 99.0 \\
5.6 Luna Frame (medium) & \best{\ci{18.2}{14.8}{22.3}} & \ci{63.7}{58.9}{68.3} & \ci{+45.5}{39.8}{51.5} & 100.0 \\
\bottomrule
\end{tabular}}
\caption{Reasoning mode. Strict pair accuracy on all 400 pairs of the indicated axis under matched prompts.
Reasoning effort is an API setting; differences use a paired bootstrap and
accuracy cells show 95\% Wilson intervals.}
\label{tab:reasoning-production}
\end{table}

\begin{table}[H]
\centering
\scriptsize
\setlength{\tabcolsep}{3pt}
\begin{tabular}{lrrr}
\toprule
Model & T & B & Cumulative \\
\midrule
DeepSeek V4 Flash & \shortstack{50.9\\{\scriptsize 98.8/3.1}} & \shortstack{52.5\\{\scriptsize 96.2/8.8}} & \shortstack{23.1\\{\scriptsize 44.8/1.5}} \\
DeepSeek V4 Pro & \shortstack{50.9\\{\scriptsize 98.8/3.1}} & \shortstack{49.4\\{\scriptsize 98.8/0.0}} & \shortstack{53.0\\{\scriptsize 97.8/8.3}} \\
GPT-5.6 Luna & \shortstack{62.2\\{\scriptsize 100.0/24.4}} & \shortstack{47.5\\{\scriptsize 86.2/8.8}} & \shortstack{69.2\\{\scriptsize 64.9/73.5}} \\
GPT-5.6 Terra & \shortstack{64.1\\{\scriptsize 100.0/28.1}} & \shortstack{50.6\\{\scriptsize 100.0/1.2}} & \shortstack{92.1\\{\scriptsize 99.3/84.8}} \\
Claude Sonnet 5 & \best{\shortstack{88.1\\{\scriptsize 100.0/76.2}}} & \best{\shortstack{58.1\\{\scriptsize 93.8/22.5}}} & \best{\shortstack{98.9\\{\scriptsize 100.0/97.7}}} \\
\bottomrule
\end{tabular}
\caption{Label-specific formula discrimination. Within-condition balanced
accuracy is shown with True/False accuracy below each estimate. Malformed
outputs count as incorrect.}
\label{tab:within-condition-detail}
\end{table}

\begin{table}[H]
\centering
\scriptsize
\setlength{\tabcolsep}{2.5pt}
\begin{tabular}{lrrrrr}
\toprule
Model & Reported & Parsed & Parsed $n$ & Malf. & Empty/other \\
\midrule
DeepSeek V4 Flash & 11.6 & 12.1 & 770 & 30 & 30/0 \\
DeepSeek V4 Pro & 18.6 & 18.6 & 800 & 0 & 0/0 \\
GPT-5.6 Luna & 38.2 & 38.2 & 800 & 0 & 0/0 \\
GPT-5.6 Terra & 62.9 & 62.9 & 800 & 0 & 0/0 \\
Claude Sonnet 5 & \best{79.1} & \best{90.3} & 701 & 101 & 98/3 \\
\bottomrule
\end{tabular}
\caption{Output-format sensitivity. Overall strict accuracy with malformed responses counted wrong
(Reported) and conditional on both sides parsing (Parsed). The final columns
count malformed sides and split empty from nonempty responses. Parsed accuracy
is diagnostic and does not replace the reported metric.}
\label{tab:parse-diagnostics}
\end{table}

\begin{table}[H]
\centering
\scriptsize
\setlength{\tabcolsep}{2.6pt}
\begin{tabular}{lrrrr}
\toprule
& \multicolumn{2}{c}{Frame} & \multicolumn{2}{c}{Domain} \\
Model & Prem. & Conj. & Prem. & Conj. \\
\midrule
DeepSeek V4 Flash & 11.8 & 0.0 & 21.5 & 13.5 \\
DeepSeek V4 Pro & 0.5 & 11.0 & 36.0 & 27.0 \\
GPT-5.6 Luna & 31.3 & 5.5 & 55.0 & 61.5 \\
GPT-5.6 Terra & 37.4 & 23.5 & \best{96.5} & \best{93.0} \\
Claude Sonnet 5 & \best{80.5} & \best{61.0} & 92.5 & 82.0 \\
\bottomrule
\end{tabular}
\caption{Premise-ablation performance. Strict accuracy is grouped by ablation
status. The retained counts are
Frame: 195 premise-dependent, 200 conjecture-only, and 5 unresolved; Domain: 200 premise-dependent and 200 conjecture-only.}
\label{tab:premise-analysis}
\end{table}

\begin{table}[H]
\centering
\scriptsize
\setlength{\tabcolsep}{1pt}
\begin{tabular}{lrrrrr}
\toprule
Model & Consistent & \shortstack{Per-contrast\\$n$} & Best fit & Macro agr. & \shortstack{Fit\\support} \\
\midrule
DeepSeek V4 Flash & 172/400 & 16--49 & K & \ci{91.3}{86.1}{95.7} & 100.0 \\
DeepSeek V4 Pro & 212/400 & 10--71 & T & \ci{78.4}{72.8}{84.7} & 61.1 \\
GPT-5.6 Luna & 370/400 & 67--79 & T & \ci{84.7}{81.7}{87.8} & 100.0 \\
GPT-5.6 Terra & 347/400 & 55--80 & T & \ci{85.1}{81.7}{88.5} & 100.0 \\
Claude Sonnet 5 & 357/400 & 64--80 & K & \best{\ci{95.3}{93.1}{97.3}} & 100.0 \\
\bottomrule
\end{tabular}
\caption{Semantic-affinity diagnostics. Per-contrast $n$ is the range across five
contrasts; best-fit support and macro-agreement intervals use a
contrast-stratified bootstrap. Pro's 61.1\% support indicates an unstable fit.}
\label{tab:implicit-logic-detail}
\end{table}

\begin{table}[H]
\centering
\scriptsize
\setlength{\tabcolsep}{0.45pt}
\resizebox{\columnwidth}{!}{\begin{tabular}{lrrrrrr}
\toprule
Model & Fit & $n$ & Acc. T & Acc. F & $\Delta_Y\mid T$ & $\Delta_Y\mid F$ \\
\midrule
DeepSeek V4 Flash & K & 172 & \ci{97.1}{93.4}{98.8} & \ci{12.2}{8.1}{17.9} & -- & -- \\
DeepSeek V4 Pro & T & 212 & \ci{99.1}{96.6}{99.7} & \ci{4.7}{2.6}{8.5} & \ci{-2.1}{-8.1}{+1.7} & \ci{-10.7}{-22.0}{-0.3} \\
GPT-5.6 Luna & T & 370 & \ci{97.0}{94.8}{98.3} & \ci{20.3}{16.5}{24.7} & \ci{+4.9}{+2.2}{+8.0} & \ci{+0.7}{-7.9}{+8.8} \\
GPT-5.6 Terra & T & 347 & \best{\ci{99.7}{98.4}{99.9}} & \ci{32.3}{27.6}{37.4} & \ci{+0.5}{+0.0}{+1.4} & \ci{-28.4}{-38.2}{-18.2} \\
Claude Sonnet 5 & K & 357 & \ci{98.6}{96.8}{99.4} & \best{\ci{71.4}{66.5}{75.9}} & -- & -- \\
\bottomrule
\end{tabular}}
\caption{Bias-controlled affinity transfer. The table links semantic affinity to explicit judgments,
using only polarity-consistent items. $\Delta_Y\mid y$ is the explicit Yes-rate
difference between contrasts the fitted profile predicts True versus False,
within gold label $y$; positive values are predicted by a fixed-profile account.
Accuracy cells show Wilson intervals and differences show bootstrap intervals.
For K fits the profile never predicts True here, so the effect is unidentified
(--), rather than conflated with label bias.}
\label{tab:affinity-prediction}
\end{table}

\begin{table}[H]
\centering
\scriptsize
\setlength{\tabcolsep}{3.5pt}
\begin{tabular}{lrrr}
\toprule
Model & Named & Defined & TPTP \\
\midrule
GPT-4.1 & \ci{4.0}{1.1}{13.5} & \best{\ci{10.0}{4.3}{21.4}} & \ci{0.0}{0.0}{7.1} \\
DeepSeek V4 Flash & \ci{6.0}{2.1}{16.2} & \best{\ci{10.0}{4.3}{21.4}} & \ci{24.0}{14.3}{37.4} \\
GPT-5.6 Terra & \best{\ci{38.0}{25.9}{51.8}} & \ci{6.0}{2.1}{16.2} & \best{\ci{44.0}{31.2}{57.7}} \\
\bottomrule
\end{tabular}
\caption{Representation sensitivity. Strict pair accuracy on Frame under matched representations.
Definitional English spells out relational conditions without property names;
formal TPTP presents the complete problem in non-classical TPTP. Intervals are
95\% Wilson intervals. GPT-4.1 parses at least 97.0\%
of sides in every representation; DeepSeek's named/definitional/formal parse
rates are 91.0/89.0/94.0\%. GPT-4.1 and the unversioned Flash endpoint are retained from
an earlier pilot; Terra is the main-evaluation endpoint.}
\label{tab:representation}
\end{table}

\begin{table}[H]
\centering
\scriptsize
\setlength{\tabcolsep}{2.8pt}
\begin{tabular}{lrrrrrr}
\toprule
Contrast & $n$ & Val./NLI & Modal & Quant. & Nodes & Dual \\
\midrule
B/S5 & 80 & 40/40 & 2 & 0 & 8 & 50.0 \\
K/D & 80 & 40/40 & 1 & 0 & 8 & 50.0 \\
K/T & 80 & 40/40 & 1 & 0 & 8 & 50.0 \\
T/B & 80 & 40/40 & 2 & 0 & 8 & 50.0 \\
T/S4 & 80 & 40/40 & 2 & 0 & 8 & 43.8 \\
cumulative/constant & 132 & 66/66 & 1 & 1 & 8 & 47.3 \\
varying/cumulative & 134 & 67/67 & 1 & 1 & 8 & 39.6 \\
varying/decreasing & 134 & 67/67 & 1 & 1 & 8 & 39.6 \\
\bottomrule
\end{tabular}
\caption{Oracle and structural audit. Coverage is outcome-asymmetric: all
800 False sides and only 73 True sides are single-prover resolutions, so raw model accuracy
by resolution is label-confounded rather than an independent oracle check.
Modal and Quant. are median depths, Nodes is median conjecture AST size, and
Dual is the percentage of sides with agreement between both provers. Pair sides
have identical complexity by construction.}
\label{tab:benchmark-audit}
\end{table}

\begin{table}[H]
\centering
\scriptsize
\setlength{\tabcolsep}{3pt}
\begin{tabular}{lrrrr}
\toprule
Contrast & Families & Skeletons & Premise skel. & Largest \\
\midrule
B/S5 & 2 & 61 & 33 & 4 \\
K/D & 2 & 64 & 30 & 3 \\
K/T & 2 & 64 & 33 & 3 \\
T/B & 2 & 61 & 33 & 5 \\
T/S4 & 2 & 66 & 37 & 3 \\
cumulative/constant & 2 & 93 & 47 & 5 \\
varying/cumulative & 2 & 90 & 44 & 4 \\
varying/decreasing & 2 & 86 & 40 & 3 \\
\bottomrule
\end{tabular}
\caption{Formula-skeleton diversity. Counts follow replacement of every proposition or predicate
with a placeholder. A generator family is a contrast--mode--modal-schema cell;
Premise skel. counts NLI skeletons and Largest is the largest substitution
cluster. Main-table intervals resample these skeleton clusters.}
\label{tab:schema-audit}
\end{table}

\section{Reasoning and Prompt Protocols}

We call inference with reasoning mode disabled \emph{direct}.
Production reasoning calls retain that prompt and temperature zero.
OpenRouter routing and sampling parameters otherwise use their defaults;
temperature zero, maximum output tokens, reasoning fields, timeouts, and client
concurrency are the explicitly recorded exceptions.
Flash uses reasoning mode with \texttt{reasoning.effort: high} and maximum output
tokens set to 4,096; Luna uses \texttt{reasoning.effort: medium} with 2,048.
The full balanced-core Flash rerun uses 512 tokens direct and 8,192 at high
effort, preventing hidden deliberation from exhausting the token budget before
the final binary answer.
Returned hidden reasoning, token usage, exact request parameters, endpoint,
timestamp, and cost are archived for every response.
The parser accepts only case-insensitive Yes/No with optional terminal
punctuation, or a final \texttt{Answer: Yes/No} marker; it performs no repair.

We retain an earlier matched GPT-4.1 pilot because it isolates a prompted
rationale from reasoning mode. Its direct template ends with
\emph{Answer only Yes or No}. The rationale condition instead appends:
\emph{Reason step by step about how the
stated semantic conditions affect the inference. In at most five sentences,
without headings or restating the problem, give a concise derivation and end
with exactly `Answer: Yes' or `Answer: No'.} The DeepSeek condition retains the
direct prompt and sends \texttt{reasoning.effort: high} with a 2,048-token
maximum output tokens. Thus the latter changes inference-time computation without
adding reasoning language to the prompt.

\paragraph{Matched qualitative example.}
For a T--K pair, direct DeepSeek answers Yes/Yes, while high effort gives Yes/No, matching the oracle. On the K side its final reasoning is: \emph{If there are no accessible worlds, then the premise is vacuously true. Then the conjecture might be false at w0. So it's possible that the premise is true but the conjecture is false. Therefore, the answer is No.}

\paragraph{Persistent rationale error.}
On a T--S4 pair, the T-side oracle is False, but GPT-4.1's rationale ends: \emph{Under reflexivity, if $\phi$ is true at all accessible worlds from w, then at any accessible world v (including w itself), $\phi$ is also true at all worlds accessible from v, because v is accessible from itself. Therefore, the statement is valid under the given semantic specification.} The argument incorrectly treats reflexivity as if it propagated accessibility paths; that step requires transitivity.

\begin{table}[H]
\centering
\scriptsize
\setlength{\tabcolsep}{2pt}
\begin{tabular}{lrrrr}
\toprule
Model & Direct & Reasoning & $\Delta$ & Parse \\
\midrule
GPT-4.1 & \best{\ci{6.0}{2.1}{16.2}} & \ci{30.0}{19.1}{43.8} & \ci{+24}{+12}{+36} & 100.0 \\
DeepSeek V4 Flash & \best{\ci{6.0}{2.1}{16.2}} & \best{\ci{86.0}{73.8}{93.0}} & \ci{+80}{+66}{+92} & 94.0 \\
\bottomrule
\end{tabular}
\caption{Prompted rationale versus reasoning mode pilot. Strict pair accuracy on the matched 50-pair Frame mini-set.
Accuracy cells show 95\% Wilson intervals; Parse is the enhanced-condition
pair parse rate, and differences use a paired bootstrap. GPT-4.1 is an earlier
pilot endpoint using a concise rationale prompt; DeepSeek uses high-effort
reasoning mode.}
\label{tab:reasoning-mini}
\end{table}

\paragraph{Direct controlled-English template.}
\begin{quote}\small
\texttt{Semantic specification:}\\
\texttt{- [frame condition]}\\
\texttt{- [domain condition, when relevant]}\\[2pt]
\texttt{Premises:}\\
\texttt{1. [local premise]}\\
\texttt{Conjecture: [conjecture]}\\[2pt]
\texttt{Question: Does the conjecture follow from the premises under this
semantic specification?}\\
\texttt{Answer only Yes or No.}
\end{quote}
For validity items, \emph{Premises} and \emph{Conjecture} are replaced by
\emph{Statement}, and the question asks whether that statement is valid.

\end{document}